\documentclass[aps,prb,reprint,superscriptaddress,amsmath,amssymb]{revtex4-2}

\usepackage{graphicx}
\usepackage{bm}
\usepackage{hyperref}
\usepackage{caption} 

\usepackage{float}
\usepackage{amsmath}
\usepackage{mathtools}

\hypersetup{
    colorlinks=true,
    linkcolor=blue,
    filecolor=magenta,      
    urlcolor=blue,
    citecolor=blue,
}

\makeatletter
\renewcommand{\fnum@figure}{Figure~\thefigure:}
\makeatother

\begin{document}

\title{Probability-Preserving Transformer for the Time-Dependent Schrödinger Equation}
\author{Mushtaq Ali}
\email{mushtaq.ali@ue.edu.pk}
\affiliation{Department of Physics, Division of Science and Technology, University of Education Lahore, Pakistan}
\author{Muzamil Tariq}
\affiliation{Department of Physics, Division of Science and Technology, University of Education Lahore, Pakistan}
\author{Niaz Ali Khan}
\email{niazkhan@xmu.edu.cn}
\affiliation{Department of Physics, Xiamen University, Xiamen 361005, China}

\begin{abstract}
Solving the time-dependent Schrödinger equation (TDSE) via traditional numerical methods is computationally intensive. Transformer models offer a compelling alternative, but standard implementations rely on soft constraints that cannot rigorously guarantee probability conservation. Here, we introduce a Transformer architecture that enforces TDSE probability conservation as a hard constraint. The design intrinsically ensures unitarity across temporal evolution without requiring repeated retraining. Our empirical findings show that this hard-constraint approach is not only physically exact but also computationally superior to conventional soft-constraint methods.
\end{abstract}
\maketitle
\section{Introduction}
The Schrödinger equation is the cornerstone of non-relativistic quantum mechanics, describing the evolution of quantum states and enabling the prediction of observables such as position, momentum, and energy \cite{griffiths2018introduction,sakurai2020modern,shankar2012principles}. The wavefunction $\psi(x,t)$, obtained by solving the Schrödinger equation, encodes the complete probabilistic description of a quantum system. Its squared modulus, $|\psi(x,t)|^2$, is interpreted as the probability density of finding the particle at position $x$ and time $t$. The dynamical evolution of quantum states is described by the time-dependent Schrödinger equation (TDSE), which governs how the wavefunction evolves over time \cite{griffiths2018introduction, sakurai2020modern, shankar2012principles}. The TDSE is expressed as
\begin{equation}
i\hbar \frac{\partial \psi(x,t)}{\partial t} = \hat{H} \psi(x,t),
\end{equation}
where $\psi(x,t)$ is the complex-valued wavefunction, $\hbar$ is the reduced Planck constant, and $\hat{H}$ is the Hamiltonian operator representing the total energy of the system \cite{griffiths2018introduction, sakurai2020modern}. By governing the evolution of quantum systems under diverse physical conditions, the TDSE provides the theoretical backbone for a wide spectrum of complex quantum dynamics, including non-equilibrium quench dynamics, many-body simulations, and ultrafast phenomena \cite{griffiths2018introduction, sakurai2020modern, shankar2012principles, Niaz2024PRE, Niaz2024CPC, fauseweh2024quantum, wolter2024ultrafast}. Furthermore, to ensure that the particle exists somewhere within the physical domain, the TDSE inherently conserves total probability through the fundamental normalization condition \cite{griffiths2018introduction, sakurai2020modern, shankar2012principles, chin2024symplectic}:
\begin{equation}
\int |\psi(x,t)|^2 dx = 1.\label{eq:normalizationConditions}
\end{equation}

The TDSE is the fundamental equation governing the time evolution of quantum systems \cite{griffiths2018introduction, sakurai2020modern, shankar2012principles, feit1982solution, chin2024symplectic, fauseweh2024quantum}. It is essential for understanding the dynamics of cavity quantum electrodynamics (cQED), which describes the interaction between quantum emitters and electromagnetic fields \cite{blais2021circuit}. By predicting how quantum states evolve in time, the TDSE provides insights into phenomena such as Rabi oscillations \cite{kockum2019ultrastrong} and entanglement dynamics \cite{bluvstein2024logical}. It lies at the core of ultrafast electron dynamics and laser–matter interactions \cite{heide2024petahertz}. In semiconductor physics and nanoelectronics, the TDSE rigorously models quantum tunneling and electron transport, informing device design and next-generation materials \cite{purves2024quantum}. Moreover, its intersection with artificial intelligence has spurred the development of operator learning frameworks \cite{li2020fourier, lu2021learning, kovachki2023neural}, Transformer-based quantum simulators \cite{geneva2022transformers, sprague2024transformer, cha2024attention}, and efficient surrogate models \cite{pfau2020abinitio, hermann2020deep, ma2024quantum}.

Numerical solvers or machine learning architectures approximating TDSE solutions must enforce normalization [Eq.~(\ref{eq:normalizationConditions})] throughout temporal evolution to maintain physical consistency. Since analytical solutions are intractable for complex potentials, numerical discretization is essential. The most widely used techniques include the Finite Difference Method (FDM)~\cite{leveque2007finite}, which is simple and efficient but limited by stability constraints; the Crank--Nicolson method (CNM)~\cite{crank1947practical}, an implicit norm-conserving scheme requiring sparse linear solvers; the finite element method (FEM)~\cite{zienkiewicz2013finite, pepper2017finite}, which uses local basis functions over unstructured meshes but incurs high complexity and memory costs; and spectral methods \cite{feit1982solution, Niaz2021, Niaz2024CPC, Niaz2024CSF, Niaz2024PRE}, which employ global basis functions (e.g., Fourier, Chebyshev) but are sensitive to discontinuities. Collectively, for large-scale or real-time applications, these numerical solvers suffer from operational bottlenecks, such as recomputing solutions from scratch for each simulation, resulting in significant computational latency. Machine learning offers a transformative alternative by learning direct operator mappings from initial states to temporal trajectories \cite{li2020fourier, lu2021learning, kovachki2023neural}, bypassing step-by-step integration and enabling millisecond-scale inference. However, conventional neural networks lack implicit physical constraints and often violate fundamental conservation laws. This motivates the use of physics-informed neural architectures \cite{raissi2018physics, dong2021exact, son2024sobolev}, which embed exact conservation laws—especially probability conservation—into the network design to ensure physical consistency alongside fast prediction.

We introduce a probability-preserving transformer (PPT) solution for the TDSE---a framework that reconciles deep learning speed with physical exactness by embedding probability conservation as an intrinsic architectural constraint, rather than a soft penalty \cite{dong2021exact}. The PPT guarantees physical validity by explicitly normalizing the predicted complex wavefunction at every forward pass, without sacrificing the expressive power of self-attention for quantum dynamics \cite{vaswani2017attention, geneva2022transformers, sprague2024transformer}. A multi-token temporal representation enhances the model's ability to capture phase transformations while ensuring exact conservation. We benchmark the PPT against standard Transformers and the Crank–Nicolson solver \cite{vaswani2017attention, cao2021choose, crank1947practical} using metrics of accuracy, conservation, and efficiency—demonstrating its potential as a physically consistent and computationally efficient alternative for quantum evolution.

The remainder of this paper is structured as follows. Section~\ref{sec:NumericalMethod} presents the numerical framework for solving the TDSE for the 1D infinite potential well. Section~\ref{sec:Numericalimplementation} details the implementation of the PPT, with a focus on the multi-token variant. Section~\ref{sec:NumericalPerformance} examines the computational performance of the proposed approach. Finally, the last section summarizes our conclusions.
\section{Numerical Methods\label{sec:NumericalMethod}}
We begin with a 1D infinite potential well as an ideal benchmark for evaluating the core TDSE physics \cite{griffiths2018introduction,sakurai2020modern, shankar2012principles}. The TDSE governs the temporal evolution of the wavefunction, with the current quantum state determining its subsequent evolution, thereby naturally forming a sequential process \cite{griffiths2018introduction, sakurai2020modern, shankar2012principles, chin2024symplectic}. This inherent sequentiality makes the TDSE uniquely well-suited for Transformer architectures, whose self-attention mechanisms are specifically designed to model complex temporal dependencies across consecutive steps \cite{vaswani2017attention, cao2021choose, geneva2022transformers, sprague2024transformer}. The TDSE is
\begin{equation}
i\hbar \frac{\partial \psi(x,t)}{\partial t} = -\frac{\hbar^2}{2m} \frac{\partial^2 \psi(x,t)}{\partial x^2} + V(x)\psi(x,t),
\end{equation}
where $V(x)$ denotes the potential energy of the infinite square well, defined as
\begin{equation}
V(x) = \begin{cases} 
0, & 0 < x < L \\
\infty, & \text{otherwise.} 
\end{cases}
\end{equation}
Within the well, the potential is zero $(V(x) = 0)$, and the Schrödinger equation simplifies to
\begin{equation}
i\hbar \frac{\partial \psi(x,t)}{\partial t} = -\frac{\hbar^2}{2m} \frac{\partial^2 \psi(x,t)}{\partial x^2}.\label{eq:TDSE-V0}
\end{equation}
For the sake of simplicity, we adopt natural units, setting $\hbar = 1, m = 1, L = 1$ \cite{griffiths2018introduction,feit1982solution}. The system is subject to Dirichlet boundary conditions $\psi(0,t) = 0, \psi(1,t) = 0$ and the initial state is chosen as the ground state $\psi(x,0) = \sqrt{2} \sin(\pi x)$.
\subsection{Numerical Data Generation and Preprocessing}
To generate our dataset for this problem, we employ the CNM \cite{crank1947practical} to produce high-fidelity ground-truth reference solutions. The generated data are verified to ensure that the total probability strictly integrates to unity \cite{chin2024symplectic, dong2021exact}, guaranteeing that the model is trained exclusively on physically consistent data. Each data row comprises a spatial coordinate, a temporal coordinate, and the corresponding real and imaginary parts of the wavefunction. The spatial grid spacing is defined as
\begin{equation}
\Delta x = \frac{L}{N_x - 1},
\end{equation}
where $N_x$ denotes the total number of spatial grid points. The spatial domain is uniformly discretized into 100 grid points, while the temporal domain is partitioned into 200 steps with a constant increment of $\Delta t=0.001$. The system is initialized using the theoretical ground-state solution \cite{griffiths2018introduction, sakurai2020modern} to satisfy both normalization and boundary constraints, $\psi(x,0) = \sqrt{2} \sin(\pi x)$. Additionally, homogeneous Dirichlet boundary conditions are enforced to confine the particle within the high-potential walls, reducing the system to an implicit matrix equation.
\begin{equation}
A\psi(x,t_{n+1}) = B\psi(x,t_n),
\end{equation}
where $\psi(x,t_n)$ denotes the wavefunction evaluated at the time $t_n$ and $\psi(x,t_{n+1})$ is its counterpart at the subsequent time step, for every point in the spatial discretization. Here, $A$ and $B$ are tridiagonal matrices of the form:
\begin{align}
A=& \text{tridiag}\left( -r, 1+2r, -r \right) \in \mathbb{R}^{N_x\times N_x}, \nonumber \\
B=& \text{tridiag}\left( r, 1-2r, r \right) \in \mathbb{R}^{N_x\times N_x},\quad r=\frac{i\hbar\Delta t}{4m\Delta x^2}.
\label{eq:vecter}
\end{align}
These matrices are determined by the spatial discretization and simulation parameters \cite{crank1947practical, leveque2007finite}. Importantly, a tridiagonal matrix algorithm solves the linear system at each step to obtain the wavefunction at the next time level \cite{thomas1949elliptic, press2007numerical}. The trapezoidal rule tracks the total quantum probability \cite{press2007numerical, leveque2007finite}:
\begin{equation}
P(t_n) \approx \sum_{i=1}^{N_x} \left| \psi(x_i, t_n) \right|^2 \Delta x ,\label{eq:Normalization}
\end{equation}
where $\psi(x_i, t_n)$ is the complex-valued wavefunction at $i$-th spatial grid point. Prior to storage, the wavefunction is decomposed into its real and imaginary parts, resulting in a dataset entry comprising four features:
\begin{equation}
(x, t, \operatorname{Re}[\psi], \operatorname{Im}[\psi]).
\end{equation}
It is noteworthy that the Crank–Nicolson scheme does not necessitate the decomposition of the wavefunction into its real and imaginary components. Nevertheless, standard Transformer architectures are inherently designed for real vector spaces. Consequently, during data preparation, the complex wavefunction is separated into its real and imaginary parts, producing a real-valued representation without sacrificing neural network trainability and ensuring exact retention of the underlying quantum state. The real and imaginary components are subsequently recombined at inference to reconstruct the complex wavefunction.

Although point-wise data representations are adequate for conventional regression tasks, enforcing global constraints requires the Transformer architecture to process the full wavefunction at each time step. Consequently, the data are restructured accordingly, with the total probability evaluated using the trapezoidal rule as presented by Eq.~(\ref{eq:Normalization}).

Unlike the Crank–Nicolson scheme, the transformer does not inherently preserve normalization during inference; the predicted wavefunction may initially violate unit probability. To remedy this, we introduce a hard normalization layer that explicitly computes the total probability and rescales the wavefunction to enforce unit normalization:
\begin{equation}
\Psi(x_i,t_n) =
\frac{\psi(x_i,t_n)}
{\sqrt{P(t_n)}},
\end{equation}
where $\Psi(x_i,t_n)$ represents the predicted wavefunction at the $i$-th spatial grid point following the application of the probability-preserving layer. The hard normalization is performed after decomposing the complex wavefunction into its real and imaginary parts to ensure compatibility with the real-valued Transformer architecture. To enable a systematic evaluation of spatial-temporal representations, the data are organized into two distinct formats.

\subsubsection{Single-token baseline}
In the single-token baseline, each input is a single time scalar, serving as a baseline to evaluate whether conventional neural architectures can inherently preserve probability without explicit constraints \cite{raissi2018physics, dong2021exact}. The raw point-wise data are systematically transformed to reconstruct the full wavefunction. Specifically, after sorting the spatial coordinates in ascending order, the real and imaginary parts of the wavefunction are extracted at each time step $t_n$, and the resulting components are concatenated into a single feature vector.
\begin{align}
\psi(t_n) = \Bigl[ 
    & \operatorname{Re}[\psi(x_1,t_n)], \operatorname{Re}[\psi(x_2,t_n)], \dots, \operatorname{Re}[\psi(x_{100},t_n)], \nonumber \\
    & \operatorname{Im}[\psi(x_1,t_n)], \operatorname{Im}[\psi(x_2,t_n)], \dots, \operatorname{Im}[\psi(x_{100},t_n)]
\Bigr].
\label{eq:vecter}
\end{align}
For $100$ spatial grid points, the real and imaginary components form a target vector of $200$ outputs per time step. The input is a single scalar (current time), while the target is the full wavefunction \cite{raissi2018physics, geneva2022transformers}. Across $200$ time steps, the dataset dimensions are
\begin{equation}
X_{s} \in \mathbb{R}^{200 \times 1}, \quad Y_s \in \mathbb{R}^{200 \times 200}.
\end{equation}
The single-token dataset is represented by the input matrix $X_s \in \mathbb{R}^{200\times1}$ and the target matrix $Y_s \in \mathbb{R}^{200\times200}$. Here, the 200 rows correspond to the 200 discrete time steps generated by the Crank--Nicolson solver. Each row of $X_s$ contains a single time value $t_i$, representing one temporal token. In contrast, each corresponding row of $Y_s$ contains the complete wavefunction at that time step, represented by 200 values: 100 values for the real component and 100 values for the imaginary component across the spatial grid. Thus, each single input token is mapped to the complete quantum state at the corresponding time. This format supports the quantum probability-preserving layer, which computes the total probability directly from the predicted output during training and inference.

\subsubsection{Multi-token baseline}
In the multi-token representation, consecutive time steps are grouped into windows, allowing the Transformer's self-attention \cite{vaswani2017attention, geneva2022transformers, cao2021choose} to learn temporal dependencies and phase dynamics. Since the initial inputs are unit-length sequences, the self-attention mechanism cannot inherently model temporal relationships \cite{vaswani2017attention, geneva2022transformers}. Thus, dataset engineering is required to leverage the model's sequence-learning capacity \cite{vaswani2017attention, cao2021choose}. A sliding-window approach \cite{beltagy2020longformer, cao2021choose} groups five consecutive time steps as input
\begin{equation}
X = \left[t_i, t_{i+1}, t_{i+2}, t_{i+3}, t_{i+4}\right],
\end{equation}
with the target being the wavefunction at the next step:
\begin{equation}
Y = \psi(t_{i+5})
\end{equation}
This multi-token format enables the Transformer's self-attention to capture temporal dependencies across previous states—a key advantage over recurrent models \cite{vaswani2017attention, geneva2022transformers, cao2021choose}. Unlike numerical solvers that integrate the TDSE step-by-step, the Transformer predicts the future state in a single forward pass \cite{crank1947practical, li2020fourier, lu2021learning}. The resulting data dimensions are
\begin{equation}
X_{\text{m}} \in \mathbb{R}^{195 \times 5}, \quad Y_{\text{m}} \in \mathbb{R}^{195 \times 200},
\end{equation}
where $X_m$ is the input matrix ($195$ steps $\times 5$ scalar) and $Y_m$ is the target matrix ($195$ steps $\times 200$ wavefunction values) in the multi-token data representation.
\begin{figure*}[!t]
\centering
\vspace{-5pt}
\includegraphics[width=0.95\textwidth]{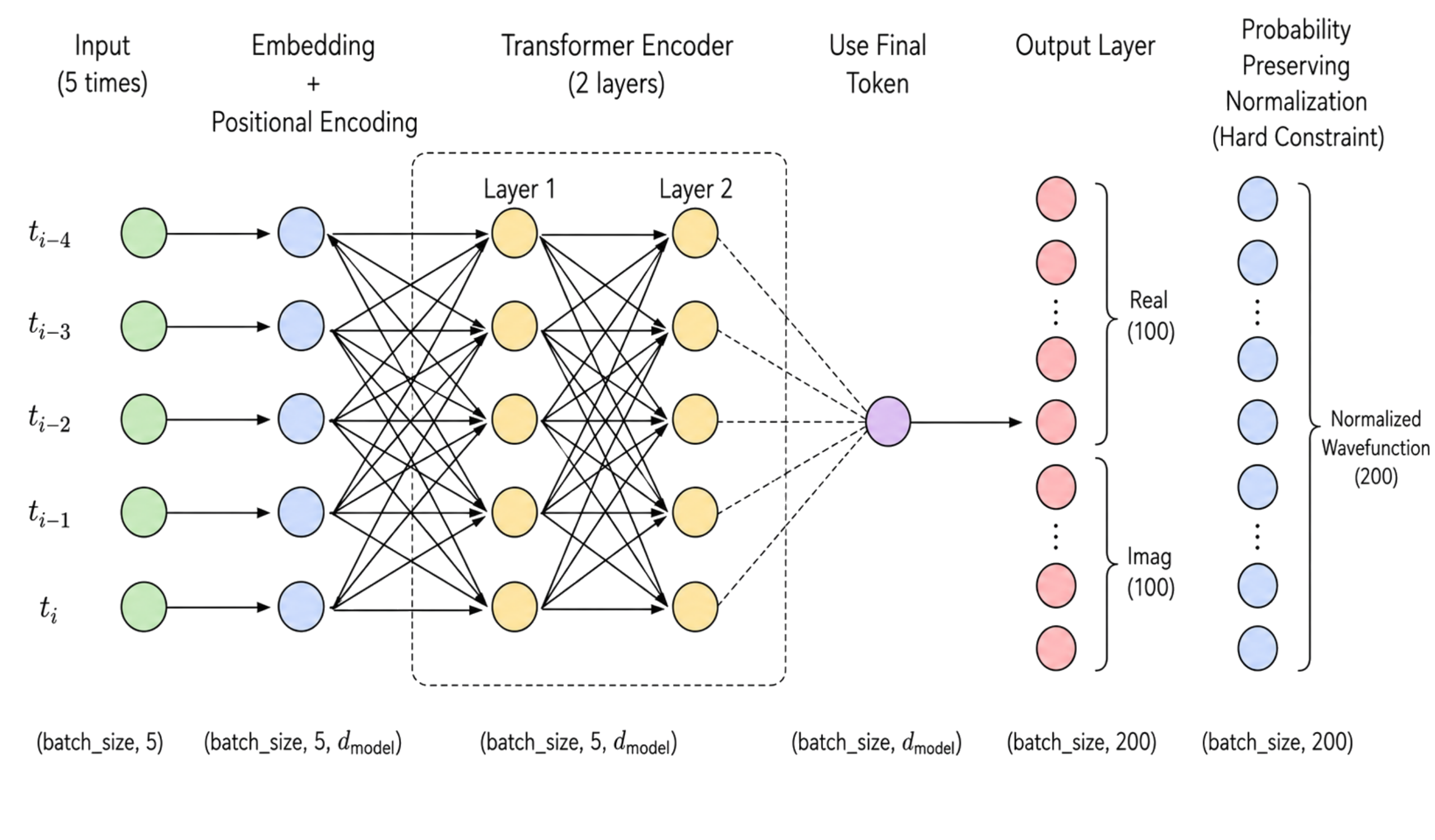}
\vspace{-8pt}
\caption{Architectural schematic of the proposed multi-token PPT. The model comprises input time embeddings, a multi-head self-attention encoder, a linear output projection, and a final hard-normalization layer designed to preserve unit probability.}
\label{fig:transformer_layer}
\vspace{-10pt}
\end{figure*}
\section{Numerical Framework and Implementation\label{sec:Numericalimplementation}}
A structured numerical framework is employed to rigorously evaluate the proposed PPT, systematically progressing through six distinct model configurations to isolate the impact of individual design choices. The evaluation begins with a baseline Multi-Layer Perceptron (MLP), which serves to assess whether the data are learnable by a neural network and whether standard architectures inherently preserve probability conservation \cite{raissi2018physics, dong2021exact}.
\begin{table}[H]
\centering
\captionsetup{}
\caption{Violation of the probability conservation in the MLP Technique.}
\label{tab:mlp_validation}
\vspace{4pt}
\renewcommand{\arraystretch}{1.4}
\begin{tabular}{l @{\hspace{25pt}} c}
\noalign{\hrule height 1.5pt}
\noalign{\vspace{5pt}}
\textbf{Test} & \textbf{\shortstack[c]{Total Probability ($P$)}} \\
\noalign{\vspace{4pt}}
\noalign{\hrule height 1pt}
\noalign{\vspace{5pt}}
Sample 0 & 1.039418 \\
Sample 1 & 1.041417 \\
Sample 2 & 1.043455 \\
Sample 3 & 1.045530 \\
\noalign{\vspace{4pt}}
\noalign{\hrule height 1.2pt}
\end{tabular}
\end{table}

Although the MLP baseline successfully reproduces the general quantum dynamics, it does not preserve the predicted total probability norm for the wavefunction \cite{raissi2018physics, griffiths2018introduction, sakurai2020modern}. As shown in Table~I, the predicted total probability ranges from $1.039418$ to $1.045530$, consistently exceeding the required normalized value of $P=1$. The progressive increase across the evaluated samples further indicates that the deviation is systematic rather than a small random fluctuation. This behavior is attributable to the absence of an explicit architectural mechanism for enforcing the normalization constraint \cite{raissi2018physics, dong2021exact}. Therefore, minimizing the prediction loss alone does not guarantee preservation of the underlying physical constraint, motivating the incorporation of a hard probability-preserving mechanism into the proposed architecture \cite{dong2021exact}.
\vspace{1pt}
\begin{figure}[H]
\centering
\includegraphics[width=\linewidth]{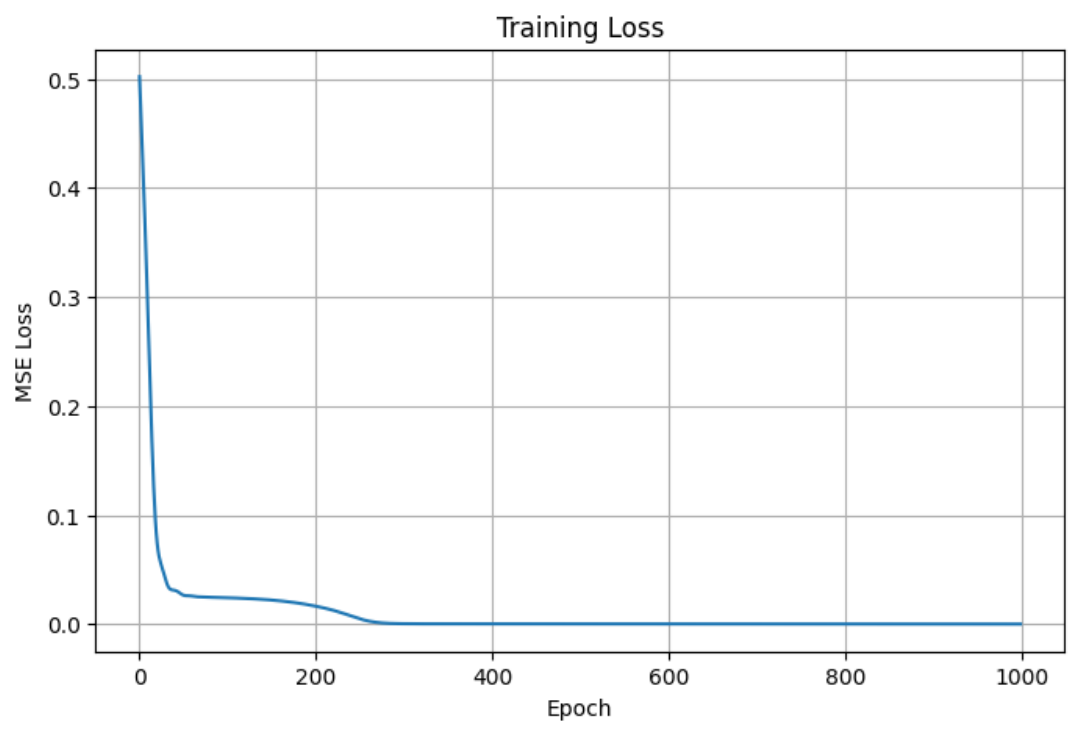}
\vspace{-14pt}
\caption{Mean squared error (MSE) of the MLP as a function of training epochs, shown over $1,000$ iterations.}
\label{fig:training_loss}
\end{figure}
\vspace{1pt}
The convergence behavior and training dynamics of the MLP baseline network over  $1{,}000$ epochs are presented in Fig.~\ref{fig:training_loss}. A rapid decrease in the loss function—from approximately $0.5$ to below $0.05$ within the initial $50$ epochs---demonstrates that the model efficiently captures the dominant spatial features and essential geometric structure of the quantum wavefunction. Having achieved a robust and well-converged state, the model is fully prepared for downstream inference tasks.

To remedy the normalization violation observed in the baseline MLP, we incorporate a hard normalization layer \cite{dong2021exact, raissi2018physics}. The network outputs the real and imaginary components of the wavefunction across all $N_x$ spatial grid points, represented as
\[
\operatorname{Re}[\psi]
=
[\hat{r}_1,\hat{r}_2,\dots,\hat{r}_{N_x}],
\qquad
\operatorname{Im}[\psi]
=
[\hat{i}_1,\hat{i}_2,\dots,\hat{i}_{N_x}]
.\]
The total probability of the unnormalized predicted wavefunction is then computed numerically over the spatial grid with spacing $\Delta x$ \cite{raissi2018physics}:
{\setlength{\abovedisplayskip}{2pt}%
\setlength{\belowdisplayskip}{2pt}%
\setlength{\abovedisplayshortskip}{2pt}%
\setlength{\belowdisplayshortskip}{2pt}%
\begin{equation}
P =
\sum_{i=1}^{N_x}
\left(
\operatorname{Re}[\psi_i]^2
+
\operatorname{Im}[\psi_i]^2
\right)\Delta x,
\end{equation}}%

\noindent where $N_x$ is the number of spatial grid points and $\Delta x$ is the spatial grid spacing. After computing the total probability, a normalization factor is derived as:
{\setlength{\abovedisplayskip}{3pt}
\setlength{\belowdisplayskip}{3pt}
\begin{equation}
\alpha = \sqrt{P + \varepsilon}.
\end{equation}}%

\noindent Here, $\varepsilon = 10^{-12}$ is a small positive constant introduced for numerical stability to prevent division by zero. Its magnitude is negligible relative to the scale of the wavefunction and therefore has a negligible effect on the normalized predictions. The real and imaginary parts of the unnormalized predicted wavefunction are then divided by $\alpha$:
\begin{equation}
\operatorname{Re}[\Psi]
=
\frac{\operatorname{Re}[\psi]}{\alpha},
\quad
\operatorname{Im}[\Psi]
=
\frac{\operatorname{Im}[\psi]}{\alpha}.
\end{equation}

\noindent To construct the final valid wavefunction, these real and imaginary components are concatenated again:
{\setlength{\abovedisplayskip}{3pt}
\setlength{\belowdisplayskip}{3pt}
\begin{equation}
\boldsymbol{\Psi} =
\left[
\operatorname{Re}[\Psi],
\operatorname{Im}[\Psi]
\right].
\end{equation}
\noindent The normalized wavefunction is substituted back into the probability expression:
{\setlength{\abovedisplayskip}{3pt}
\setlength{\belowdisplayskip}{3pt}
\begin{equation}
\begin{aligned}
\sum_{i=1}^{N_x} & \left[ \left( \frac{\operatorname{Re}[\psi_i]}{\alpha} \right)^2 + \left( \frac{\operatorname{Im}[\psi_i]}{\alpha} \right)^2 \right] \Delta x \\
& = \frac{1}{\alpha^2} \sum_{i=1}^{N} \left( \operatorname{Re}[\psi_i]^2 + \operatorname{Im}[\psi_i]^2 \right) \Delta x.
\end{aligned}
\end{equation}}

\noindent Since $\alpha^2 \approx P$ (neglecting the small stability constant $\varepsilon$), the quantum normalization condition is automatically satisfied for every predicted wavefunction independent of raw network outputs \cite{dong2021exact}. Unlike standard physics-informed loss formulations, no additional penalty-based hyperparameters are required to conserve probability \cite{raissi2018physics, karniadakis2021physics}. We are able to preserve probability conservation up to numerical precision. This successful validation motivated us to implement this hard-constraint normalization layer \cite{dong2021exact} into both single-token and multi-token Transformer architectures \cite{vaswani2017attention, geneva2022transformers, cao2021choose}. 

\subsection{Multi-token PPT Architecture}
As illustrated in Fig.~\ref{fig:transformer_layer}, the proposed Multi-token PPT processes a sequence of temporal inputs and maps them to a complete quantum state through four main stages: temporal embedding, positional encoding, Transformer-based temporal feature extraction, and constrained state reconstruction. The overall architecture follows the sequence-processing framework of Transformer models while adapting it to the temporal evolution of physical systems \cite{vaswani2017attention, geneva2022transformers}.
\subsubsection{Input processing and temporal embedding}
The model receives a sequence of five consecutive time values, $[t_{i-4},t_{i-3},t_{i-2},t_{i-1},t_i]$, corresponding to an input tensor of shape $(\text{batch\_size},5)$. Each scalar time value is independently projected into a $d_{\mathrm{model}}=64$ dimensional representation using a linear embedding layer. Learnable positional embeddings are then added to retain the ordering of the temporal tokens, which is essential for distinguishing the relative positions of observations within the input sequence \cite{vaswani2017attention}. The resulting representation has the shape $(\text{batch\_size},5,64)$.

\subsubsection{Transformer encoder}

The embedded sequence is processed by two stacked Transformer encoder layers. Each layer employs multi-head self-attention to model relationships among the five temporal tokens, enabling the network to extract temporal information from the preceding sequence \cite{vaswani2017attention}. The use of self-attention for modeling temporal dependencies is also consistent with Transformer-based approaches developed for physical systems and partial differential equations \cite{geneva2022transformers}. The sequence and embedding dimensions remain unchanged throughout the encoder, resulting in an output representation of shape $(\text{batch\_size},5,64)$.

\subsubsection{Quantum state reconstruction}

After temporal feature extraction, the representation corresponding to the final input token is selected as the summary of the preceding temporal sequence. This operation reduces the encoded representation to $(\text{batch\_size},64)$. A fully connected output layer then projects this latent representation into a 200-dimensional state vector. The resulting vector contains 100 components representing the real part and 100 components representing the imaginary part of the predicted wavefunction: $[\psi_{\mathrm{real}}, \psi_{\mathrm{imag}}]$. This representation provides the complete spatial state required for subsequent enforcement of the global probability constraint.

\subsubsection{Hard probability-preserving constraint}

The unconstrained output is subsequently passed through the hard probability-preserving normalization layer described in the preceding section. Rather than imposing probability conservation through an additional penalty term in the loss function, the physical constraint is incorporated directly into the network architecture. Such architectural enforcement of physical constraints is consistent with approaches that impose conservation laws and boundary conditions explicitly within neural-network formulations \cite{dong2021exact}.

The normalization layer operates on the complete predicted quantum state and transforms the unconstrained output into a normalized wavefunction satisfying the required probability condition. The final output therefore retains the same 200-dimensional representation, consisting of the real and imaginary components of the normalized wavefunction at all spatial grid points. Consequently, the proposed architecture combines temporal sequence modeling through self-attention with explicit enforcement of the quantum probability constraint, providing a Transformer-based framework for physically constrained wavefunction prediction \cite{vaswani2017attention, geneva2022transformers, dong2021exact}.

\subsection{Numerical Validation}

\begin{table}[H]
\centering
\caption{Performance comparison between the classical CNM and the proposed multi-token PPT architecture.}
\label{tab:cnm_vs_ppt_comparison}
\vspace{4pt}
\resizebox{\linewidth}{!}{%
\renewcommand{\arraystretch}{1.25}
\begin{tabular}{lcc}
\hline\hline
\textbf{Performance Metric} & \textbf{CNM} & \textbf{Multi-token PPT} \\
\hline
Minimum Total Probability ($P_{\min}$) & $1.000000000000$ & $0.999999821186$ \\
Maximum Total Probability ($P_{\max}$) & $1.000000000006$ & $1.000000238419$ \\
Mean Total Probability ($\bar{P}$)     & $1.000000000000$ & $1.000000000000$ \\
Probability Deviation / Std. Dev. ($\sigma$) & $5.4800 \times 10^{-14}$ & $1.0324 \times 10^{-7}$ \\
Mean Squared Error (MSE)              & $7.1230 \times 10^{-9}$ & $5.8011 \times 10^{-4}$ \\
Root Mean Squared Error (RMSE)        & $8.4400 \times 10^{-5}$ & $2.4086 \times 10^{-2}$ \\
Maximum Absolute Error                & $1.1990 \times 10^{-4}$ & $7.4499 \times 10^{-2}$ \\
\hline\hline
\end{tabular}%
}
\end{table}

The final probability-preserving multi-token Transformer strictly maintains the quantum probability normalization condition while providing a highly accurate learned approximation of the wavefunction evolution. This result demonstrates the benefit of incorporating physical constraints directly into the architecture, rather than relying exclusively on the optimization objective to preserve physical consistency \cite{raissi2018physics, dong2021exact, geneva2022transformers}. Unlike the Crank–Nicolson method, which obtains the wavefunction through sequential numerical time integration, the proposed PPT learns the temporal evolution from generated reference data and predicts the complete quantum state in a single forward inference pass.

The multi-token PPT achieves a mean squared error (MSE) of $5.8011 \times 10^{-4}$ and a root mean squared error (RMSE) of $2.4086\times 10^{-2}$, indicating a relatively small average deviation between the predicted wavefunctions and the corresponding Crank–Nicolson reference solutions. The maximum absolute error of $7.4499\times 10^{-2}$ represents the largest pointwise deviation observed across the evaluated wavefunction components. These errors should be interpreted as the approximation error of the learned surrogate rather than as a direct measure of the numerical accuracy of the CNM, which serves as the reference solution \cite{crank1947practical, leveque2007finite}. The difference in error magnitude is therefore expected, as CNM and PPT perform fundamentally different computational tasks.

The probability results demonstrate that the proposed architecture successfully preserves the required physical normalization. Across the evaluated predictions, the minimum and maximum probabilities are $0.999999821186$ and $1.000000238419$, respectively, while the mean probability remains $1$. The standard deviation of $1.0324 \times 10^{-7}$ further indicates that the predicted probability remains tightly concentrated around unity throughout the evaluation. This behavior is consistent with the normalization requirement of quantum mechanics, where the integral of the probability density over the physical domain must remain unity \cite{griffiths2018introduction, sakurai2020modern, shankar2012principles}.

Importantly, the comparison underscores that CNM and PPT are complementary in terms of numerical accuracy. CNM yields highly accurate reference trajectories, with an MSE of $7.1230 \times 10^{-9}$ and a probability deviation of $5.48 \times 10^{-14}$. By contrast, PPT provides a learned approximation with significantly reduced computational demands, while strictly enforcing probability conservation. These findings support the use of PPT as a physically constrained surrogate for efficient quantum-state prediction, rather than as a substitute for the high-fidelity numerical solver employed to generate reference data \cite{crank1947practical, leveque2007finite, raissi2018physics}.
\begin{figure}[htbp]
    \centering
    \includegraphics[width=\linewidth]{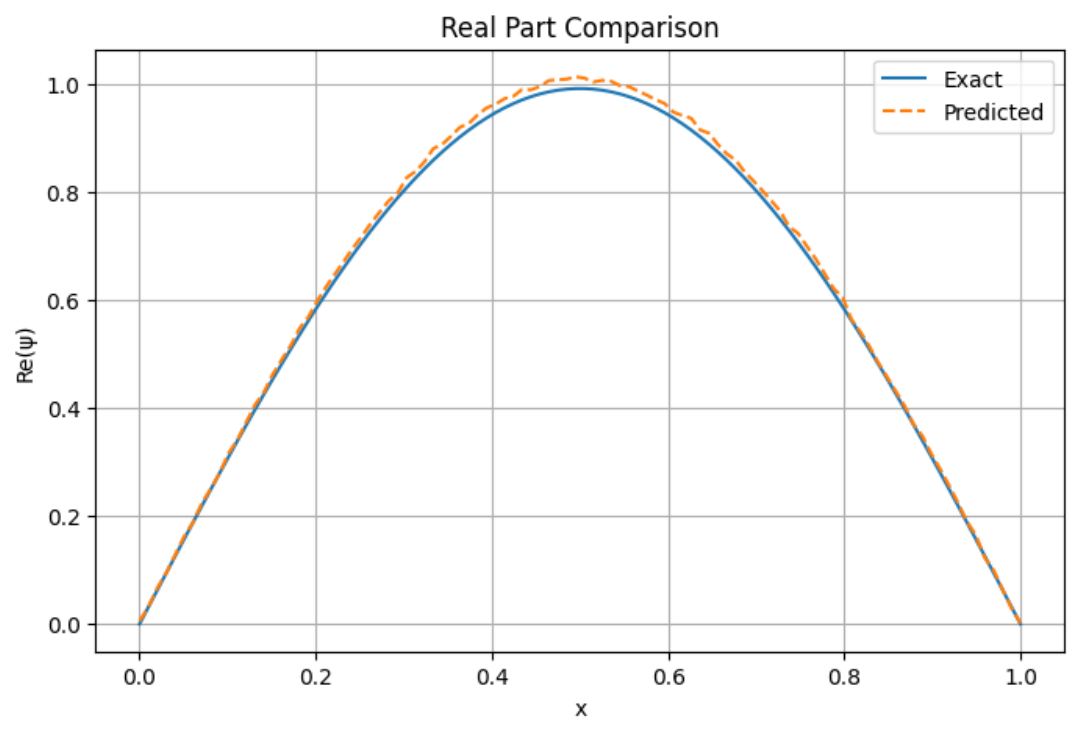}
    
    \vspace{1em} 
    
    \includegraphics[width=\linewidth]{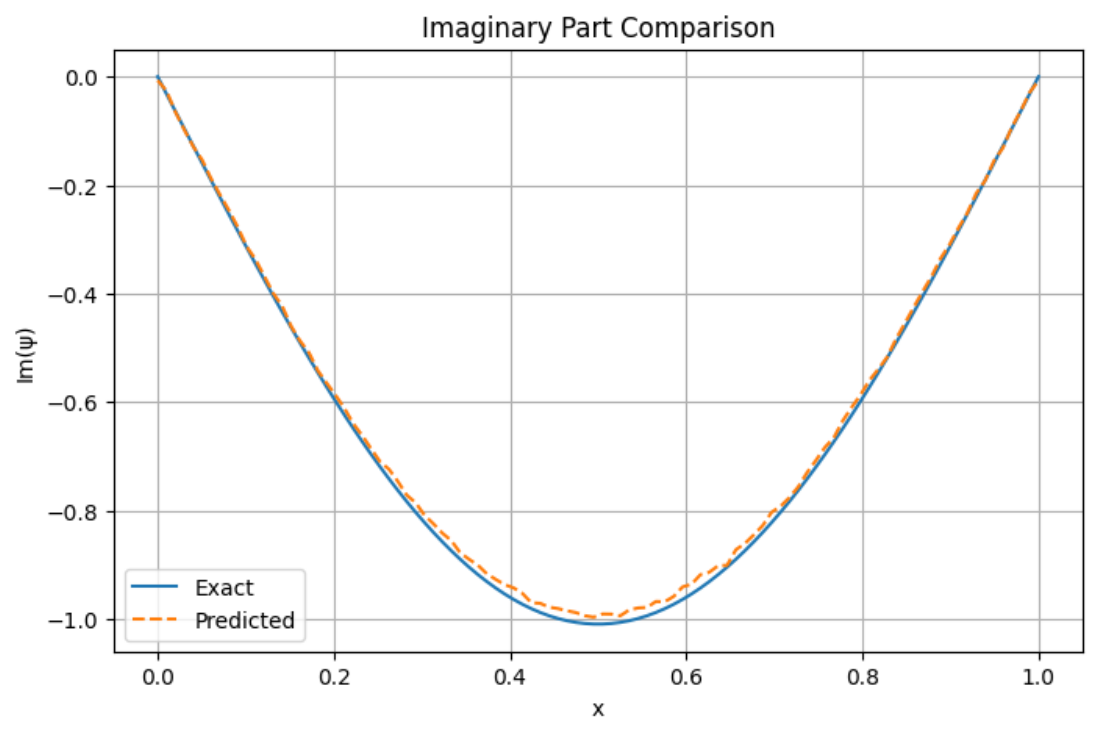}
    
    \caption{Comparison of the real ($\text{Re}(\psi)$, top) and imaginary ($\text{Im}(\psi)$, bottom) components across the spatial domain.}
    \label{fig:wavefunction_components}
\end{figure}

The model precisely captures the quantum system behavior, as shown by both the real and imaginary components in Fig.~\ref{fig:wavefunction_components} \cite{geneva2022transformers, dong2021exact}. It successfully reproduces the complex phase rotation ($e^{-iEt/\hbar}$) \cite{griffiths2018introduction, sakurai2020modern}, with the wavefunction vanishing at the boundaries of the infinite well \cite{griffiths2018introduction}, and preserves the $\pi/2$ phase relationship between components. The residual $2\%$ to $3\%$ error suggests the need for Sobolev training to regularize derivatives for future energy and momentum calculations \cite{czarnecki2017sobolev, raissi2018physics}. This is because the current normalization layer enforces only total probability conservation, without explicitly regularizing spatial smoothness \cite{dong2021exact, raissi2018physics}.
\begin{figure}[htbp]
\centering
\includegraphics[width=\linewidth]{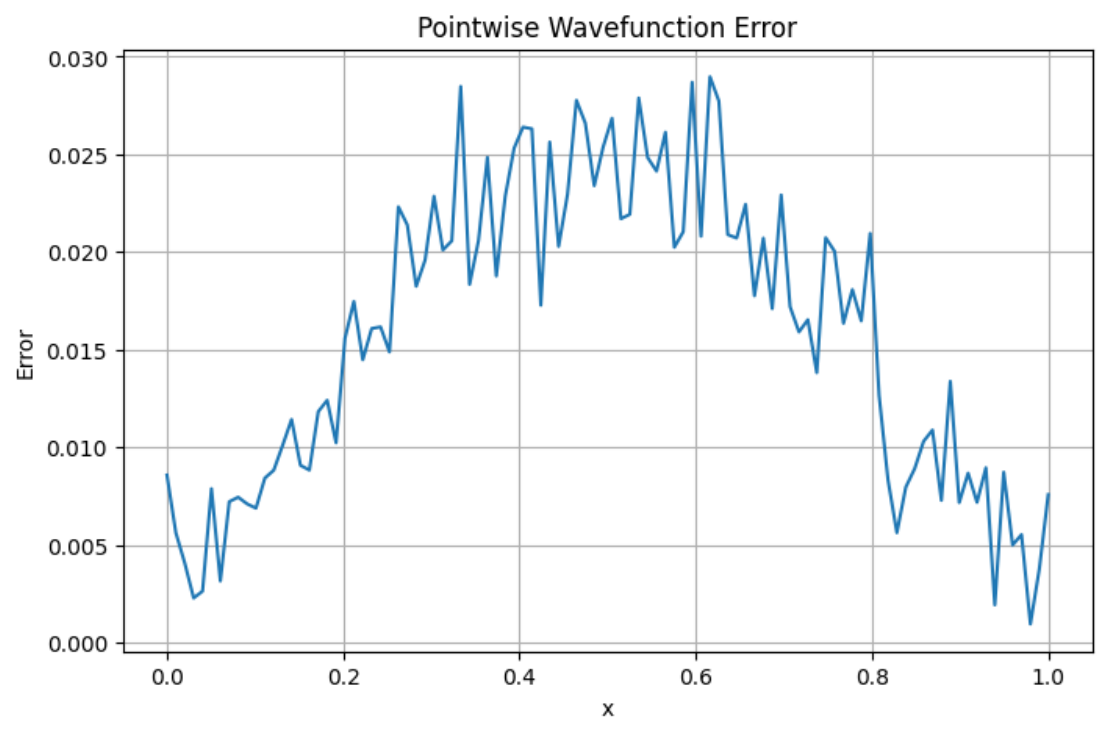}
\vspace{-8pt}
\caption{Pointwise wavefunction error across the spatial domain grid points.}
\label{fig:pointwise_error}
\vspace{-6pt}
\end{figure}

The pointwise wavefunction error quantifies the local discrepancy between the predicted and reference wave functions at each spatial location. Across spatial coordinates it is computed as the Euclidean distance between their respective real and imaginary components:
{\setlength{\abovedisplayskip}{3pt}
\setlength{\belowdisplayskip}{3pt}
\begin{equation}
\begin{aligned}
E(x,t) = \Big[ &\left(\text{Re}(\psi_{\text{pred}}) - \text{Re}(\psi_{\text{true}})\right)^2 + \\
               &\left(\text{Im}(\psi_{\text{pred}}) - \text{Im}(\psi_{\text{true}})\right)^2 \Big]^{1/2}.
\end{aligned}
\end{equation}}

\noindent This metric reveals the distribution of prediction error across space and identifies regions where the model deviates most from the reference solutions. As illustrated in Figure~\ref{fig:pointwise_error} a peak relative error of roughly $2\%$ to $3\%$ represents a highly accurate prediction.

\section{Computational Performance\label{sec:NumericalPerformance}}

To evaluate the computational efficiency of the proposed PPT, we benchmark its inference time against the classical CN numerical solver \cite{crank1947practical, leveque2007finite}. The CNM, being a sequential integrator, requires the computation of every intermediate time step to predict the wavefunction at a future horizon \cite{press2007numerical, leveque2007finite}. By contrast, the trained PPT directly maps historical input to future states in a single forward pass, thereby circumventing iterative time stepping \cite{vaswani2017attention, geneva2022transformers, cao2021choose, li2020fourier}. Both approaches were tested under identical spatial resolution:
{\setlength{\abovedisplayskip}{3pt}
\setlength{\belowdisplayskip}{3pt}
\begin{equation}
N_x = 100 \quad \text{and} \quad \Delta t = 0.001.
\end{equation}}%

For each horizon, we recorded three runtime metrics: (1) Sequential CNM Runtime—the time for sequential integration over all intermediate steps \cite{crank1947practical, leveque2007finite}; (2) PPT Batched—parallel throughput for multi-state inference \cite{vaswani2017attention, geneva2022transformers}; and (3) PPT Constant-Time Inference—the cost of a single future-state prediction \cite{li2020fourier, lu2021learning}. All measurements were averaged over $R=1000$ independent runs to reduce hardware jitter \cite{press2007numerical}.
{\setlength{\abovedisplayskip}{2pt}
\setlength{\belowdisplayskip}{2pt}
\begin{equation}
T = \frac{1}{R} \sum_{i=1}^{R} T_i.
\end{equation}}%

\noindent The speedup factor to represent relative performance gain is expressed via:
{\setlength{\abovedisplayskip}{2pt}
\setlength{\belowdisplayskip}{2pt}
\begin{equation}
S = \frac{T_{\text{CN}}}{T_{\text{PPT}}}.
\end{equation}}%
\vspace{1pt}

\noindent where $S > 1$ indicates superior Transformer efficiency as prediction horizon expand. As the CNM scales linearly with time, the PPT maintains a fixed inference cost. For longer temporal predictions, the computational advantage of proposed approach becomes more significant.

\begin{table}[htbp]
\centering
\captionsetup{}
\caption{Comparative computational performance of the classical CNM and the proposed multi-token PPT.}
\label{tab:runtime_comparison}
\vspace{4pt}
\renewcommand{\arraystretch}{1.4}
\setlength{\tabcolsep}{6pt}
\resizebox{\linewidth}{!}{%
\begin{tabular}{ccccc}
\noalign{\hrule height 1.2pt}
\noalign{\vspace{5pt}}
\textbf{\shortstack[c]{Prediction\\Horizon ($N$)}} & \textbf{\shortstack[c]{Crank--Nicolson\\Sequential (s)}} & \textbf{\shortstack[c]{PPT\\Batched (s)}} & \textbf{\shortstack[c]{PPT\\Constant-Time (s)}} & \textbf{Speedup} \\
\noalign{\vspace{4pt}}
\noalign{\hrule height 0.6pt}
\noalign{\vspace{5pt}}
10  & 0.000612 & 0.000897 & 0.000579 & 1.06$\times$  \\
20  & 0.003159 & 0.001589 & 0.000636 & 4.96$\times$  \\
50  & 0.003158 & 0.002209 & 0.000561 & 5.63$\times$  \\
100 & 0.009180 & 0.003832 & 0.000589 & 15.58$\times$ \\
150 & 0.012253 & 0.006075 & 0.000928 & 13.20$\times$ \\
195 & 0.015476 & 0.007106 & 0.000600 & 25.78$\times$ \\
\noalign{\vspace{4pt}}
\noalign{\hrule height 1.2pt}
\end{tabular}%
}
\end{table}
To ensure a fair comparison, both the CNM solver and the PPT model were benchmarked on the same CPU hardware under identical computational conditions. For each prediction horizon, the runtime was measured over 1,000 independent repetitions, and the reported values represent the average runtime to reduce the effect of transient hardware and system-level timing fluctuations \cite{press2007numerical}. The evaluation reveals a stark contrast in computational scaling. The CNM, which propagates iteratively from the base condition, exhibits linear runtime growth with the prediction horizon—from $0.000612s$ at $N=10$ to $0.01576s$ at $N=195$—reflecting the cumulative cost of sequential time stepping \cite{crank1947practical, leveque2007finite, press2007numerical}. In sharp contrast, the PPT model achieves constant-time inference, with runtime remaining between $0.00056s$ to $0.00093s$ irrespective of how far forward the prediction extends \cite{vaswani2017attention, geneva2022transformers, cao2021choose, li2020fourier}.
\begin{figure}[H]
\centering
\begin{minipage}{\linewidth}
    \centering
    \includegraphics[width=0.88\linewidth]{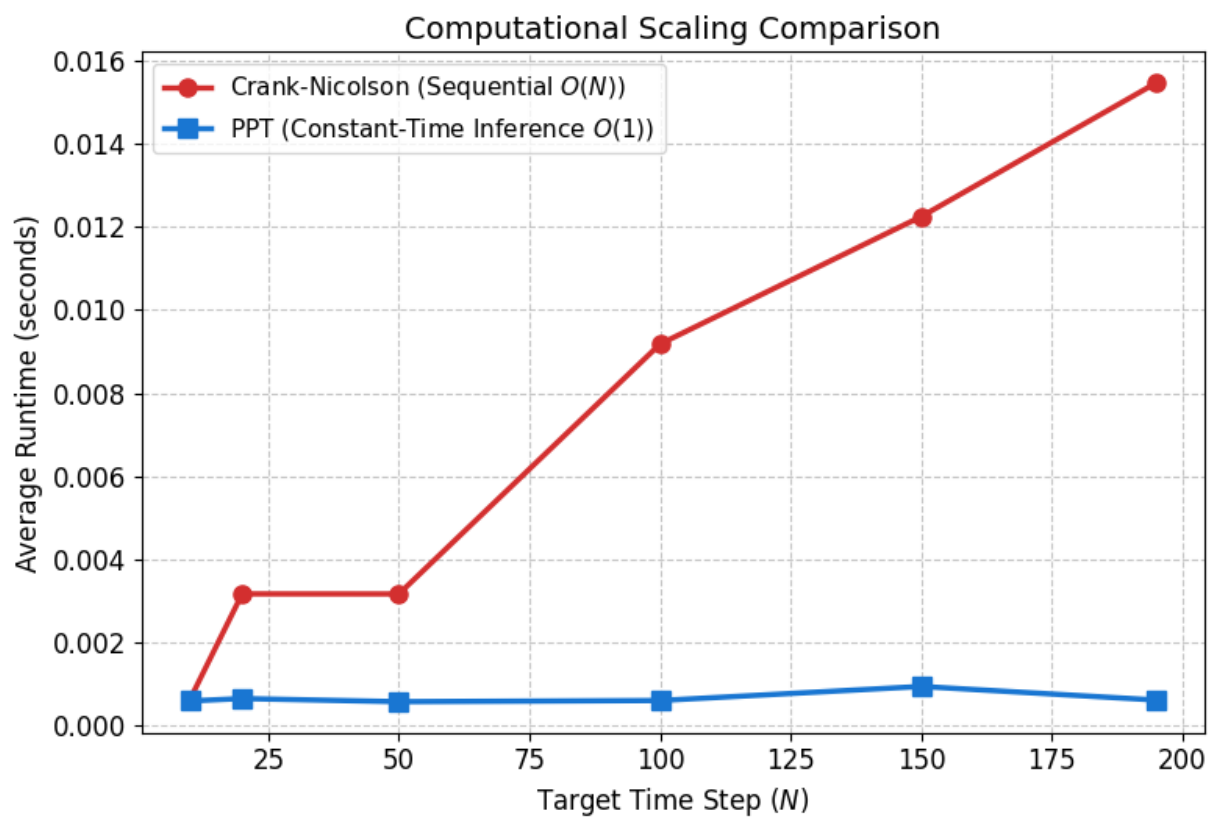}
    \vspace{2pt}
    \centerline{\small (a) Computational Scaling Comparison}
\end{minipage}
\vspace{1.5em}
\begin{minipage}{\linewidth}
    \centering
    \includegraphics[width=0.88\linewidth]{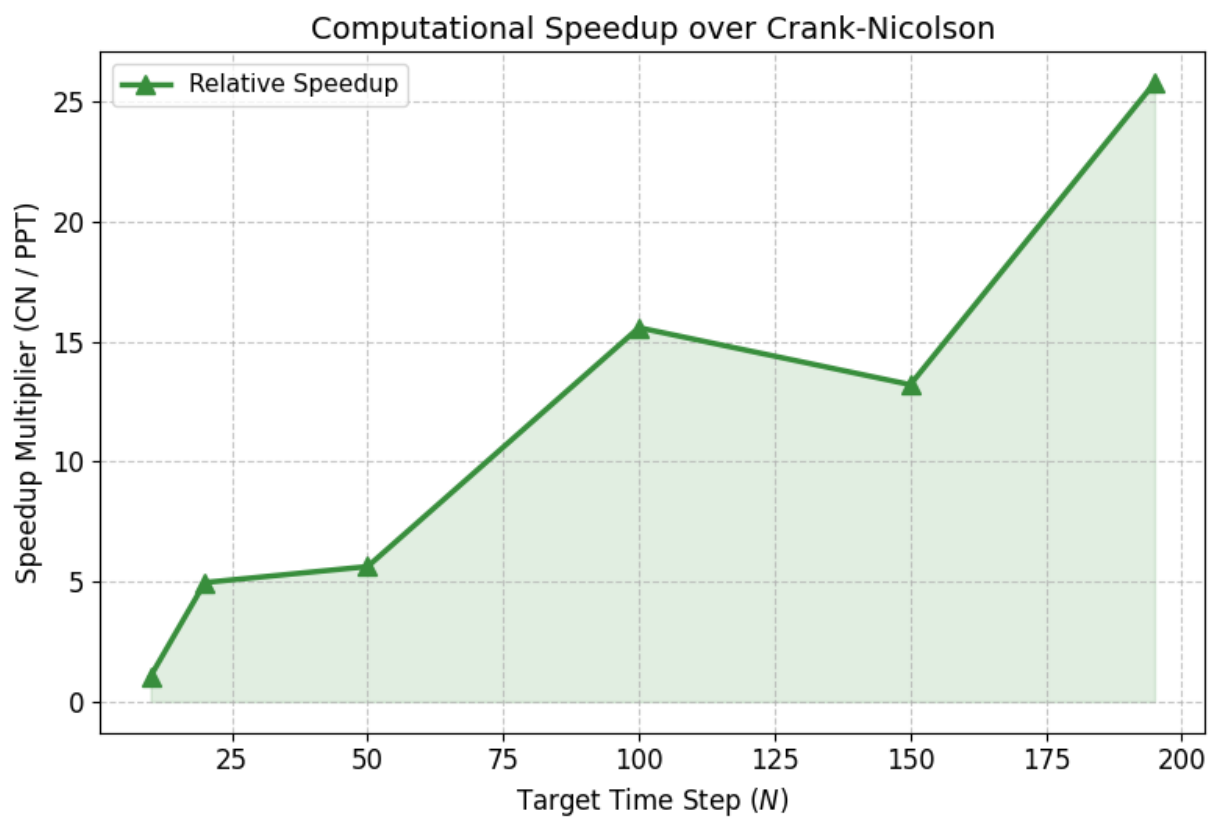}
    \vspace{2pt}
    \centerline{\small (b) Relative Speedup over Crank--Nicolson}
\end{minipage}

\vspace{6pt}
\caption{Computational scaling and relative performance speedup of the proposed PPT framework versus the sequential Crank--Nicolson solver.}
\label{fig:computational_performance}
\end{figure}
\vspace{-2pt}
As the prediction horizon extends, the numerical solver exhibits an approximately linear increase in execution time (red curve), since the Crank-Nicolson method advances the solution sequentially through the required intermediate time steps \cite{crank1947practical, leveque2007finite, press2007numerical}. In contrast, the PPT maintains a nearly constant single-state inference time of approximately $0.0006\,\text{s}$ across the tested horizons because the trained model predicts the future wavefunction directly from the temporal input sequence, bypassing explicit intermediate steps \cite{vaswani2017attention, geneva2022transformers, cao2021choose}. With the use of Transformer-based surrogate models, this distinction is consistent for predicting dynamical physical systems, where the trained network replaces repeated numerical propagation during inference \cite{geneva2022transformers, li2020fourier, lu2021learning}. Consequently, the relative computational advantage of the PPT becomes more pronounced for longer prediction horizons, as the computational cost of sequential CNM propagation continues to accumulate while the cost of an individual PPT forward pass remains approximately horizon-independent within the tested range \cite{crank1947practical, leveque2007finite, geneva2022transformers}. These results demonstrate the potential of the proposed framework as an efficient learned surrogate for repeated quantum wavefunction prediction, while CNM remains the numerical method used to generate the reference states \cite{geneva2022transformers, vaswani2017attention}.
\vspace{6pt}
\section{Conclusion\label{sec:Conclusion}}
This study demonstrated the effectiveness of the Probability-Preserving Transformer for solving the 1D time-dependent Schrödinger equation while enforcing probability conservation via a hard-normalization layer. The results confirmed that embedding physical constraints into Transformer architectures yields physically consistent solutions. However, scaling to more complex systems requires moving beyond simple normalization to Sobolev training, which penalized derivative discrepancies essential for accurate momentum and kinetic energy calculations \cite{czarnecki2017sobolev, raissi2018physics}. Extension to higher dimensions was enabled by patch-based tokenization \cite{cao2021choose, dosovitskiy2021image}, while predicting trajectories across an entire quantum family from a single model required the inclusion of energy-conditional tokens \cite{vaswani2017attention, peurifoy2018nanophotonic}.
\section*{DATA AVAILABILITY}
The source code, trained models, and datasets used in this study are publicly available in \href{https://github.com/Muzamil-Tariq-resolver/Probability-Preserving-Transformer-TDSE}{Repository}

\bibliography{references}

@article{vaswani2017attention,
  title={Attention is all you need},
  author={Vaswani, Ashish and Shazeer, Noam and Parmar, Niki and Uszkoreit, Jakob and Jones, Llion and Gomez, Aidan N and Kaiser, {\L}ukasz and Polosukhin, Illia},
  journal={Advances in Neural Information Processing Systems},
  volume={30},
  pages={5998--6008},
  year={2017},
  url={https://proceedings.neurips.cc/paper/2017/hash/3f5ee243547dee91fbd053c1c4a845aa-Abstract.html}
}

@article{raissi2018physics,
  title={Physics-informed neural networks: A deep learning framework for solving forward and inverse problems involving nonlinear partial differential equations},
  author={Raissi, Maziar and Perdikaris, Paris and Karniadakis, George Em},
  journal={Journal of Computational Physics},
  volume={378},
  pages={686--707},
  year={2019},
  publisher={Elsevier},
  doi={10.1016/j.jcp.2018.10.045},
  url={https://doi.org/10.1016/j.jcp.2018.10.045}
}

@book{griffiths2018introduction,
  title={Introduction to Quantum Mechanics},
  author={Griffiths, David J and Schroeter, Darrell F},
  edition={3rd},
  year={2018},
  publisher={Cambridge University Press},
  doi={10.1017/9781108575705},
  url={https://doi.org/10.1017/9781108575705}
}

@book{sakurai2020modern,
  title={Modern Quantum Mechanics},
  author={Sakurai, Jun John and Napolitano, Jim},
  edition={3rd},
  year={2020},
  publisher={Cambridge University Press},
  doi={10.1017/9781108587280},
  url={https://doi.org/10.1017/9781108587280}
}

@book{shankar2012principles,
  title={Principles of Quantum Mechanics},
  author={Shankar, Ramamurti},
  edition={2nd},
  year={1994},
  publisher={Springer Science \& Business Media},
  doi={10.1007/978-1-4757-0576-8},
  url={https://doi.org/10.1007/978-1-4757-0576-8}
}

@article{crank1947practical,
  title={A practical method for numerical evaluation of solutions of partial differential equations of the heat-conduction type},
  author={Crank, John and Nicolson, Phyllis},
  journal={Mathematical Proceedings of the Cambridge Philosophical Society},
  volume={43},
  number={1},
  pages={50--67},
  year={1947},
  publisher={Cambridge University Press},
  doi={10.1017/S0305004100023197},
  url={https://doi.org/10.1017/S0305004100023197}
}

@article{feit1982solution,
  title={Solution of the Schr{\"o}dinger equation by a spectral method},
  author={Feit, M D and Fleck Jr, J A and Steiger, A},
  journal={Journal of Computational Physics},
  volume={47},
  number={3},
  pages={412--433},
  year={1982},
  publisher={Elsevier},
  doi={10.1016/0021-9991(82)90091-2},
  url={https://doi.org/10.1016/0021-9991(82)90091-2}
}

@book{pepper2017finite,
  title={The Finite Element Method: Basic Concepts and Applications with MATLAB, MAPLE, and COMSOL},
  author={Pepper, Darrell W and Heinrich, Juan C},
  edition={3rd},
  year={2017},
  publisher={CRC Press},
  doi={10.1201/9781315372334},
  url={https://doi.org/10.1201/9781315372334}
}

@article{li2020fourier,
  title={Fourier neural operator for parametric partial differential equations},
  author={Li, Zongyi and Kovachki, Nikola and Azizzadenesheli, Kamyar and Liu, Burigede and Bhattacharya, Kaushik and Stuart, Andrew and Anandkumar, Anima},
  journal={arXiv preprint arXiv:2010.08895},
  year={2020},
  doi={10.48550/arXiv.2010.08895},
  url={https://doi.org/10.48550/arXiv.2010.08895}
}

@article{lu2021learning,
  title={Learning nonlinear operators via DeepONet based on the universal approximation theorem of operators},
  author={Lu, Lu and Jin, Pengzhan and Pang, Guofei and Zhang, Zhongqiang and Karniadakis, George Em},
  journal={Nature Machine Intelligence},
  volume={3},
  number={3},
  pages={218--229},
  year={2021},
  publisher={Nature Publishing Group UK London},
  doi={10.1038/s42256-021-00302-5},
  url={https://doi.org/10.1038/s42256-021-00302-5}
}

@article{czarnecki2017sobolev,
  title={Sobolev training for neural networks},
  author={Czarnecki, Wojciech M and Osindero, Simon and Jaderberg, Max and Swirszcz, Grzegorz and Pascanu, Razvan},
  journal={Advances in Neural Information Processing Systems},
  volume={30},
  pages={4278--4287},
  year={2017},
  url={https://proceedings.neurips.cc/paper/2017/hash/21e031e40004ff931a74288c886e0821-Abstract.html}
}

@article{cao2021choose,
  title={Choose a transformer: Fourier or galerkin},
  author={Cao, Shuhao},
  journal={Advances in Neural Information Processing Systems},
  volume={34},
  pages={24924--24940},
  year={2021},
  url={https://proceedings.neurips.cc/paper/2021/hash/d14030282d1c68e1ad404738520e5dd9-Abstract.html}
}

@article{fauseweh2024quantum,
  title={Quantum many-body simulations on digital quantum computers: State-of-the-art and future challenges},
  author={Fauseweh, Benedikt},
  journal={Nature Communications},
  volume={15},
  number={1},
  pages={2123},
  year={2024},
  publisher={Nature Publishing Group UK London},
  doi={10.1038/s41467-024-46402-9},
  url={https://doi.org/10.1038/s41467-024-46402-9}
}

@article{heide2024petahertz,
  title={Petahertz electronics},
  author={Heide, Christian and Keathley, Phillip D and Kling, Matthias F},
  journal={Nature Reviews Physics},
  volume={6},
  number={11},
  pages={648--662},
  year={2024},
  publisher={Nature Publishing Group UK London},
  doi={10.1038/s42254-024-00752-1},
  url={https://doi.org/10.1038/s42254-024-00752-1}
}

@article{ma2024quantum,
  title={Quantum embedding method with transformer neural network quantum states for strongly correlated materials},
  author={Ma, Huan and Shang, Honghui and Yang, Jinlong},
  journal={npj Computational Materials},
  volume={10},
  number={1},
  pages={220},
  year={2024},
  publisher={Nature Publishing Group UK London},
  doi={10.1038/s41524-024-01408-y},
  url={https://doi.org/10.1038/s41524-024-01408-y}
}

@book{leveque2007finite,
  title={Finite Difference Methods for Ordinary and Partial Differential Equations: Steady-State and Time-Dependent Problems},
  author={LeVeque, Randall J},
  year={2007},
  publisher={SIAM},
  doi={10.1137/1.9780898717839},
  url={https://doi.org/10.1137/1.9780898717839}
}

@book{zienkiewicz2013finite,
  title={The Finite Element Method: Its Basis and Fundamentals},
  author={Zienkiewicz, Olgierd Cecil and Taylor, Robert Leroy and Zhu, Jian Z},
  edition={7th},
  year={2013},
  publisher={Elsevier},
  doi={10.1016/C2009-0-24909-9},
  url={https://doi.org/10.1016/C2009-0-24909-9}
}

@article{Niaz2024PRE,
  title = {Chebyshev polynomial approach to Loschmidt echo: Application to quench dynamics in two-dimensional quasicrystals},
  author = {Khan, Niaz Ali and Ye, Shihao and Zhou, Ziheng and Cheng, Shujie and Xianlong, Gao},
  journal = {Phys. Rev. E},
  volume = {109},
  issue = {6},
  pages = {065311},
  numpages = {9},
  year = {2024},
  month = {Jun},
  publisher = {American Physical Society},
  doi = {10.1103/PhysRevE.109.065311},
  url = {https://link.aps.org/doi/10.1103/PhysRevE.109.065311}
}

@article{Niaz2024CSF,
title = {Fidelity susceptibility probes of dynamical quantum criticality},
journal = {Chaos, Solitons \& Fractals},
volume = {183},
pages = {114975},
year = {2024},
issn = {0960-0779},
doi = {https://doi.org/10.1016/j.chaos.2024.114975},
url = {https://www.sciencedirect.com/science/article/pii/S0960077924005277},
author = {Niaz Ali Khan}
}

@Article{Niaz2021,
  author    = {N. A. Khan and S. T. Amin},
  title     = {Probing band-center anomaly with the Kernel polynomial method},
  journal   = {Phys. Scr.},
  year      = {2021},
  volume    = {96},
  number    = {4},
  pages     = {045812},
  month     = {Feb},
  doi       = {10.1088/1402-4896/abe322},
  publisher = {{IOP} Publishing},
  url       = {https://doi.org/10.1088/1402-4896/abe322},
}

@article{Niaz2024CPC,
title = {Linear-scale simulations of quench dynamics},
journal = {Comput. Phys. Commun.},
volume = {299},
pages = {109132},
year = {2024},
issn = {0010-4655},
doi = {https://doi.org/10.1016/j.cpc.2024.109132},
url = {https://www.sciencedirect.com/science/article/pii/S0010465524000559},
author = {Niaz Ali Khan and Wen Chen and Munsif Jan and Gao Xianlong}
}

@article{blais2021circuit,
  title = {Circuit quantum electrodynamics},
  author = {Blais, Alexandre and Grimsmo, Arne L. and Girvin, S. M. and Wallraff, Andreas},
  journal = {Rev. Mod. Phys.},
  volume = {93},
  issue = {2},
  pages = {025005},
  year = {2021},
  publisher = {American Physical Society},
  doi = {10.1103/RevModPhys.93.025005},
  url = {https://doi.org/10.1103/RevModPhys.93.025005}
}

@article{kockum2019ultrastrong,
  title = {Ultrastrong coupling between light and matter},
  author = {Frisk Kockum, Anton and Miranowicz, Adam and De Liberato, Simone and Nori, Salvatore and Nori, Franco},
  journal = {Nature Reviews Physics},
  volume = {1},
  number = {1},
  pages = {19--40},
  year = {2019},
  publisher = {Nature Publishing Group},
  doi = {10.1038/s42254-018-0006-2},
  url = {https://doi.org/10.1038/s42254-018-0006-2}
}

@article{bluvstein2024logical,
  title = {A logical quantum processor based on reconfigurable atom arrays},
  author = {Bluvstein, Dolev and Evered, Simon J. and Geim, Alexandra A. and Li, Sophie H. and Zhou, Hengyun and Manovitz, Tom and Ebadi, Sepehr and Cain, Madelyn and Kalinowski, Marcin and Hangleiter, Dominik and others},
  journal = {Nature},
  volume = {626},
  number = {7997},
  pages = {58--65},
  year = {2024},
  publisher = {Nature Publishing Group UK London},
  doi = {10.1038/s41586-023-06927-3},
  url = {https://doi.org/10.1038/s41586-023-06927-3}
}

@article{purves2024quantum,
  title = {Quantum tunneling and resonant transport dynamics in nanoscale semiconductor heterostructures},
  author = {Purves, Michael A. and Zhao, Jian-Hao and Ferry, David K.},
  journal = {Phys. Rev. B},
  volume = {109},
  number = {4},
  pages = {045412},
  year = {2024},
  publisher = {American Physical Society},
  doi = {10.1103/PhysRevB.109.045412},
  url = {https://doi.org/10.1103/PhysRevB.109.045412}
}

@article{kovachki2023neural,
  title = {Neural operator: Learning maps between function spaces with applications to PDEs},
  author = {Kovachki, Nikola and Li, Zongyi and Liu, Burigede and Azizzadenesheli, Kamyar and Bhattacharya, Kaushik and Stuart, Andrew and Anandkumar, Anima},
  journal = {Journal of Machine Learning Research},
  volume = {24},
  number = {89},
  pages = {1--97},
  year = {2023},
  url = {https://www.jmlr.org/papers/v24/21-1524.html}
}

@article{dong2021exact,
  title = {Exact enforcement of boundary conditions and conservation laws in physics-informed neural networks},
  author = {Dong, Suchuan and Ni, Ni},
  journal = {Journal of Computational Physics},
  volume = {434},
  pages = {110234},
  year = {2021},
  publisher = {Elsevier},
  doi = {10.1016/j.jcp.2021.110234},
  url = {https://doi.org/10.1016/j.jcp.2021.110234}
}

@article{pfau2020abinitio,
  title = {Ab initio solution of the many-electron Schr{\"o}dinger equation with deep neural networks},
  author = {Pfau, David and Spencer, James S. and Matthews, Alexander G. D. G. and Foulkes, W. M. C.},
  journal = {Phys. Rev. Res.},
  volume = {2},
  number = {3},
  pages = {033429},
  year = {2020},
  publisher = {American Physical Society},
  doi = {10.1103/PhysRevResearch.2.033429},
  url = {https://doi.org/10.1103/PhysRevResearch.2.033429}
}

@article{hermann2020deep,
  title = {Deep-neural-network solution of the electronic Schr{\"o}dinger equation},
  author = {Hermann, Jan and Sch{\"a}tzle, Zeno and No{\'e}, Frank},
  journal = {Nature Chemistry},
  volume = {12},
  number = {10},
  pages = {891--897},
  year = {2020},
  publisher = {Nature Publishing Group},
  doi = {10.1038/s41557-020-0544-y},
  url = {https://doi.org/10.1038/s41557-020-0544-y}
}

@article{sprague2024transformer,
  title = {Transformer-based neural network quantum states for time evolution of quantum spin systems},
  author = {Sprague, Kaleb and Viterbi, Sarah and Carrasquilla, Juan},
  journal = {Phys. Rev. B},
  volume = {109},
  number = {12},
  pages = {125105},
  year = {2024},
  publisher = {American Physical Society},
  doi = {10.1103/PhysRevB.109.125105},
  url = {https://doi.org/10.1103/PhysRevB.109.125105}
}

@inproceedings{dosovitskiy2021image,
  title = {An Image is Worth 16x16 Words: Transformers for Image Recognition at Scale},
  author = {Dosovitskiy, Alexey and Beyer, Lucas and Kolesnikov, Alexander and Weissenborn, Dirk and Zhai, Xiaohua and Unterthiner, Thomas and Dehghani, Mostafa and Minderer, Matthias and Heigold, Georg and Gelly, Sylvain and Uszkoreit, Jakob and Houlsby, Neil},
  booktitle = {International Conference on Learning Representations (ICLR)},
  year = {2021},
  url = {https://openreview.net/forum?id=YwudA333S2q}
}

@article{geneva2022transformers,
  title = {Transformers for modeling physical systems and solving PDEs},
  author = {Geneva, Nicholas and Zabaras, Nicholas},
  journal = {Journal of Computational Physics},
  volume = {460},
  pages = {111135},
  year = {2022},
  publisher = {Elsevier},
  doi = {10.1016/j.jcp.2022.111135},
  url = {https://doi.org/10.1016/j.jcp.2022.111135}
}

@article{cha2024attention,
  title = {Attention-based neural quantum state tomography and unitary dynamics reconstruction},
  author = {Cha, Peter and Giles, Connor and Kim, Eun-Ah},
  journal = {Phys. Rev. Lett.},
  volume = {132},
  number = {3},
  pages = {030401},
  year = {2024},
  publisher = {American Physical Society},
  doi = {10.1103/PhysRevLett.132.030401},
  url = {https://doi.org/10.1103/PhysRevLett.132.030401}
}

@article{chin2024symplectic,
  title = {Symplectic and unitary integrators for the time-dependent Schr{\"o}dinger equation},
  author = {Chin, Siu A.},
  journal = {Phys. Rev. E},
  volume = {109},
  number = {3},
  pages = {035302},
  year = {2024},
  publisher = {American Physical Society},
  doi = {10.1103/PhysRevE.109.035302},
  url = {https://doi.org/10.1103/PhysRevE.109.035302}
}

@article{wolter2024ultrafast,
  title = {Ultrafast electron diffraction and attosecond dynamics governed by the time-dependent Schr{\"o}dinger equation},
  author = {Wolter, Benjamin and Pullen, Michael G. and Baudisch, Matthias and Biegert, Jens},
  journal = {Nature Physics},
  volume = {20},
  number = {3},
  pages = {412--425},
  year = {2024},
  publisher = {Nature Publishing Group},
  doi = {10.1038/s41567-024-02380-x},
  url = {https://doi.org/10.1038/s41567-024-02380-x}
}

@article{son2024sobolev,
  title = {Sobolev training for physics-informed neural networks with spatial and temporal gradient constraints},
  author = {Son, Hyeongseok and Cho, Seung-Woo and Hwang, Hyung Ju},
  journal = {Computer Methods in Applied Mechanics and Engineering},
  volume = {418},
  pages = {116521},
  year = {2024},
  publisher = {Elsevier},
  doi = {10.1016/j.cma.2023.116521},
  url = {https://doi.org/10.1016/j.cma.2023.116521}
}

@article{peurifoy2018nanophotonic,
  title={Nanophotonic particle design and characterization via deep learning},
  author={Peurifoy, Ryan and Shen, Yichen and Jing, Li and Yang, Yi and Cano-Renteria, Francisco and DeLacy, Brendan G and Joannopoulos, John D and Tegmark, Max and Solja{\v{c}}i{\'c}, Marin},
  journal={Science Advances},
  volume={4},
  number={6},
  pages={eaar4206},
  year={2018},
  publisher={American Association for the Advancement of Science},
  doi={10.1126/sciadv.aar4206},
  url={https://doi.org/10.1126/sciadv.aar4206}
}

@book{press2007numerical,
  title={Numerical Recipes 3rd Edition: The Art of Scientific Computing},
  author={Press, William H and Teukolsky, Saul A and Vetterling, William T and Flannery, Brian P},
  edition={3rd},
  year={2007},
  publisher={Cambridge University Press},
  isbn={9780521880688},
  url={https://www.cambridge.org/9780521880688}
}

@article{beltagy2020longformer,
  title={Longformer: The long-document transformer},
  author={Beltagy, Iz and Peters, Matthew E and Cohan, Arman},
  journal={arXiv preprint arXiv:2004.05150},
  year={2020},
  doi={10.48550/arXiv.2004.05150},
  url={https://doi.org/10.48550/arXiv.2004.05150}
}

@techreport{thomas1949elliptic,
  title={Elliptic Problems in Linear Differential Equations over a Network},
  author={Thomas, Llewellyn H},
  type={Technical Report},
  institution={Watson Scientific Computing Laboratory, Columbia University},
  address={New York},
  year={1949},
  url={https://www.columbia.edu/cu/computinghistory/thomas.html}
}

@article{karniadakis2021physics,
  title={Physics-informed machine learning},
  author={Karniadakis, George Em and Kevrekidis, Ioannis G and Lu, Lu and Perdikaris, Paris and Wang, Sifan and Yang, Liu},
  journal={Nature Reviews Physics},
  volume={3},
  number={6},
  pages={422--440},
  year={2021},
  publisher={Nature Publishing Group UK London},
  doi={10.1038/s42254-021-00314-5},
  url={https://doi.org/10.1038/s42254-021-00314-5}
}

\end{document}